\documentclass[12pt,notitlepage]{article}
\usepackage{hyperref} 
\usepackage[authoryear,round]{natbib}
\usepackage{amssymb,amsmath,multirow,graphicx}
\usepackage{extarrows}
\usepackage{amsthm}
\usepackage{MnSymbol}
\usepackage{indentfirst}
\usepackage{afterpage}
\usepackage{graphicx}
\usepackage{subfigure}
\usepackage{epsfig}
\usepackage{bm}
\usepackage{placeins}
\usepackage{booktabs}
\usepackage{xcolor}

\usepackage{epsf}    
\usepackage{setspace}  
\usepackage{ifthen}

\newcommand{\nothing}[1]{} \nothing{This allows you to make long comments}

\newcommand{\be}{\begin{equation}}
\newcommand{\ee}{\end{equation}}

\begin{document}

\title{Adopting AI Technologies without Traps:\\ A Multi-Agent Reinforcement Learning Framework}

\author{
Jitong Yang{\footnote{Program of Electrical and Computer Engineering, University of Iowa.}}\,~~
Avinash Reddy Mudireddy{\footnote{Program of Electrical and Computer Engineering, University of Iowa.}}\,~~
Ziqing Lu {\footnote{Department of Computer Science, Wake Forest University, Winston Salem, NC.}}\,~~
Huan Cai {\footnote { Department of Economics and Business, Cornell College, Mount Vernon, IA, 52314.}}\,~~
Weiyu Xu {\footnote{Department of Electrical and Computer Engineering, University of Iowa, Iowa City, IA 52242. Email: \texttt{weiyu-xu@uiowa.edu.}}}
\footnote{The authors are listed alphabetically. }}

\maketitle

\begin{abstract}
AI technologies and AI agents play increasingly important roles in human society, including in education, economics, and research. Adopting AI technologies can bring immediate, and sometimes tremendous, rewards such as through productivity boost, but they can also bring negative effects including but not limited to losses of job and training opportunities, and cognitive capability declines. Thus, it is important to investigate the optimal ways of adopting AI technologies without different traps. In this paper, we develop a general multi-agent Markov decision process (MDP) based reinforcement learning (RL or MARL) modeling framework for determining and analyzing optimal adoption of AI technologies.  This framework can model the long-term dynamic effects of AI adoption, and lead to finding the optimal AI adoption strategies. In our modeling framework, the negative effects of AI adoption are modeled by undesirable state transitions or negative rewards in the MDPs. The proposed multi-agent RL MDPs incorporate both human agents and AI agents. The key idea of our models is that incorporating AI technologies may not only increase the immediate rewards for AI agents but also alter the states of the involved human and AI agents. In particular, we apply this framework to analyzing AI offloading leading to skill erosion, to analyzing optimal human resource structures for a firm adopting AI, and to analyzing the economic equilibria in the  trading/employment among human and AI agents. Since the multi-agent MDP based RL technique is from control, machine learning and AI, this research is essentially using an AI technique to determine the optimal ways of adopting or co-existing with AI without traps.

\end{abstract}

\newpage

\section{Introduction}

AI technologies and related agents play increasingly important roles in human society, including in education, economics, and research \citep{goldfarb2019digital,abrardi2022artificial,agrawal2023similarities,kapoor2024ai,li2024more} \citep{hammond2025multi} \citet{cai2023ai} \cite{bauer2026fallback}
\cite{aouad2026human}
\cite{dasaklis2026large}
\cite{akin2026does}
\cite{kimura2026efficiency}. Adopting AI technologies can bring immediate, and sometimes tremendous, rewards such as through productivity boost, but they can also bring negative effects including but not limited to losses of job, losses of training opportunities, and cognitive capability declines. It is thus important to investigate the optimal ways of adopting AI technologies without falling into different traps.   

Besides offering immediate rewards, adopting AI technologies can additionally bring negative effects,  but these negative effects may not be immediate and can only show up in the future.  In the optimal way of adopting the AI technologies, one thus should not only care about the short-term rewards, but also pay attention to the long-term rewards of adopting AI technologies.  However, since adopting the AI technologies is a relatively new phenomenon and people have no experience in the best way of adopting AI, it is tempting to adopt AI technologies myopically in order to obtain maximum immediate rewards.  These myopic AI adoptions, due to their natures of both replacement and augmentation of human labors and thinking, can lead to very undesirable future consequences. Then we face the following question: how to optimally adopt AI technologies so that there are sustainable long-term rewards?

In this paper, we develop a general multi-agent Markov decision process (MDP) based reinforcement learning (RL or MARL) framework for determining and analyzing optimal adoption of AI technologies.  This framework can model the dynamic long-term  effects of AI adoption, and lead to finding the optimal AI adoption strategies. In our modeling framework, the negative effects of AI adoption are modeled by state transitions or negative rewards in the MDPs. The proposed multi-agent RL framework incorporates both human agents and AI agents, and can use the policies of the MDP models to model the economic equilibria among these agents. 

 The key idea of our method of studying the AI adoption is that incorporating AI technologies and AI agents may not only increase the immediate rewards for AI agents but also alter the states of the environment, the involved human and AI agents. In the economy, adopting AI may affect the job opportunities for human and AI agents,  and change wealth distributions among agents, resulting in different wealth states or employment states the agents are in.   For example, if a community delegates all the economic tasks to cheaper outside AI agents which do not contribute to the wealth growth of this community, the community can potentially lose job and growth opportunities.  
 For a firm, using AI technologies to replace junior workers can result in unhealthy workforce preparation: when faced with loss of senior workers or other future novel challenges, the firm may become dysfunctional.  For example, a firm can be tempted to use only senior workers working with AI in completing work tasks at a low total cost.  However, that will lead to the firm's dysfunctions when senior workers leave or retire, since no junior workers were trained at the entry-level positions. For K-12 and college students, using AI to do homework can give immediate rewards such as higher grades in the courses,  but may result in students falling into the state of poorly understanding subjects of study.  For mathematicians, using AI agents to directly produce AI proofs can achieve the rewards of quickly solving math conjectures or problems; but, at the same time, leads to loss of training opportunities for junior mathematicians; and also leads to possible loss of opportunities of mathematicians making novel discoveries and understandings when they take healthy detours when solving math conjectures or problems on their own. For workers, offloading tasks to AI can affect their skill levels: while they can obtain higher immediate productivity, the offloading can affect skills so that they under-perform when there are novel tasks appearing which cannot be fully handled by AI.  

 In addition, in a system composed of multiple AI agents and human agents, where the agents can all decide ways of adopting AI technologies, the joint actions can affect the states of the environments and the agents themselves. Thus the problem of optimally adopting AI technologies become a multi-agent MDP problem instead of a single-agent MDP problem.

 In this paper, we focus on three applications and derive the corresponding optimal strategies of incorporating AIs: work offloading versus cognitive skill training; AI usage and workforce development; trading between human and AI agents. In a myopic view, it seems always optimal in terms of maximizing the immediate rewards for workers to  offload tasks to AI since they can finish tasks more efficiently;  for firms to only retain senior workers who has high productivity when working with AI, since junior workers can be replaced by AI; and for human agents to purchase products/services only from AI agents, since the AI can provide better products/services at lower cost. However,  blind adoption of AI can change the states (skill level, workforce status, and wealth distribution) of the environment and humans involved, and may not always be optimal for maximizing the overall or longer-term reward.  Our multi-agent MDP formulation can provide a principled framework for determining the optimal way of adopting the AI technologies in maximizing the overall rewards while avoiding the traps.

 This paper is summarized as follows. In Section \ref{Sec:mathmodelMarkov}, we introduce the mathematical model of multi-agent RL framework for adopting AI technologies. In Section \ref{sec:AIoffloading}, we introduce the Markov decision process for AI offloading and skill usage.  In Section \ref{sec:workforcebuildup}, we introduce the Markov decision process for a firm adopting AI and preparing its labor reserve.  In Section \ref{sec:economics}, we introduce the multi-agent economic model for systems with both human and AI agents.  We conclude our paper in Section \ref{sec: conclusions}.

\section{Related Work}

Bondi and Johnson \cite{bondi2026skill} study how AI usage
affects worker skill development and the measurement of
productivity gains. They develop a dynamic model to examine
how delegating tasks to AI influences worker skills over time.
Their results show that although AI can improve short-term
productivity, excessive reliance on AI may lead to skill decline.
They also find that standard productivity measurements may
overestimate the long-term benefits of AI because they do not fully account for changes in worker skills. Caosun and Aral \cite{caosun2026augmentation} study the trade-off between short-term productivity gains from AI and long-term worker skill development. They develop a dynamic model to examine how AI usage affects worker productivity and skill over time. Their results show that although AI can improve immediate productivity, excessive reliance on AI may reduce workers’ skills and lead to lower productivity in the long run. They also find that AI adoption can increase the skill gap between high-skilled and low-skilled workers. 

\textcolor{black}{ We note that the dynamic models in \cite{bondi2026skill}\cite{caosun2026augmentation} are mainly restricted to analyzing skill erosion of workers using AI. Our MDP model is more general and can apply to other settings such as studying the impacts of adopting AI in many other aspects, including in trading and employments \citet{cai2023ai}\cite{cai2026artificial}\cite{falk2026ai}, the growth of mathematicians, and K-12 education.  In terms of studying skill erosion through AI offloading, in contrast to our paper, \cite{bondi2026skill}\cite{caosun2026augmentation} did not consider the case where the actions of the workers are directly dependent on the state of \emph{Markov} decision process (MDP), and only let the actions essentially only depend on the time variable or index. As a result, the dynamic models in \cite{bondi2026skill}\cite{caosun2026augmentation} are deterministic dynamic systems where the states and actions are deterministic functions of time,  and not \emph{Markov random processes}.  Compared with \cite{caosun2026augmentation} \cite{bondi2026skill}, our \emph{Markov} decision process (MDP) framework allows for random and richer state transitions under different actions. In other words, for our paper, the agent's actions depend not only on time index, but also on the current MDP's state, which can take different possible values even for a given time index.  In fact, many factors can be random in adopting AI. For example, the problem types that a worker needs to solve  or work at a time index may be random, depending on random outside application needs. As another example of randomness,  workers may randomly leave or join a firm due to retirement or relocation.   In summary, the MDP framework in this paper is a more general framework than the dynamic system model in \cite{caosun2026augmentation} \cite{bondi2026skill}. }

In \citet{cai2023ai} and \cite{cai2026artificial}, instead of investigating the short-term economic interplays among various AI agents, the authors looked into the long-term economic interplays between human and AI agents,  using a Markov chain stationary distribution based model. However, the methods in \citet{cai2023ai} and \cite{cai2026artificial}  are a particular, different model than a regular MDP (even though it can be thought of as an MDP model in a broader sense), where the relation between the old joint state (where a joint state for multiple agents is described by the wealth allocation among these different agents) and the new joint state is described by another Markov transition probability matrix which depends on the interaction graph structures, and this Markov transition probability matrix only describes the inherent mechanism of transitioning between these two specific states, and is different from the transition kernel in a regular MDP model.  In \citet{cai2023ai} and \cite{cai2026artificial},  actions are the ``spending patterns'' which do not depend on the state of wealth allocations and these actions produce deterministic wealth allocation results for the next time index.  Note that the purchasing actions in \citet{cai2023ai} and \cite{cai2026artificial} include employment actions, since employment actions are just firms purchasing labor from hired and to-be-hired employees. Falk and Tsoukalas \cite{falk2026ai} study how AI automation affects employment and economic welfare. They develop a competitive economic model to examine how firms make automation decisions. Their results show that although AI automation can reduce labor costs, excessive automation may lead to job losses and lower consumer demand. They also find that competition between firms can encourage excessive AI adoption, even when it reduces overall economic welfare. However,  the model in \cite{falk2026ai} is static in that it only considers one time index, and in contrast, the method in our paper is a dynamic MDP RL method involving multiple time indices.

\section{Mathematical Model}\label{Sec:mathmodelMarkov}

In the considered MDP environment, we denote the group of agents as $N$, with $|N|=n$. The following 5-element tuple can describe the finite MDP environment for the $n$ agents: $(\mathcal{S}, \{\mathcal{A}_i\}_{i=1}^n, P, \{R_i\}_{i=1}^n, T)$, where $T$ is the finite number of time steps, $\mathcal{S}$ is the common state space of all agents, $\mathcal{A}_i$ is the action space for the $i$-th agent with $|\mathcal{A}_i| = A_i$, $\mathcal{A}=\prod_{i=1}^n \mathcal{A}_i$ is the joint action space.
We let $P_{t\rightarrow t+1}:\mathcal{S}\times \mathcal{A}\times\mathcal{S}\rightarrow[0,1]$ be the state transition probability kernel under the joint actions of all agents between time steps $t$ and $t+1$ with $t=0,1,\dots T-1$. We use ${R}_{i,t}:\mathcal{S}\times\mathcal{A}\rightarrow \mathbb{R}$ to represent the reward function of the $i$-th  agent at time index $t$, with $t=1,2,\dots,T$.  We let the time index $t$ be in the range from 0 to $T$. We refer to \textbf{time step} $t$ as the time interval starting from time index $t-1$ and ending at time index $t$, where $1\leq t \leq T$.

We let $s_t$ denote the joint environment state of all agents at time step $t$, let $a_{i,t}$ denote the action the $i$-th agent takes at time index $t$ and  let $a^*_{i,t}$ denote the optimal action of the $i$-th agent at time step $t$. The joint action of all $n$ agents at time index $t$ is $\textbf{a}_t := (a_{1,t}, a_{2,t}, \dots, a_{n,t})$.
We denote $\pi^*_{i,t}$  the optimal policy of the $i$-th agent at time index $t$, and $\pi_{i,t}^*(s_{t}) = a_{i,t}^*$. 
The deterministic joint optimal policy at time index $t$ is $\bm{\pi}^*_t = \{\pi^*_{i,t}\}_{i=1}^n$, and we write $\pi_i^*=\{\pi_{i,t}^*\}_{t=0}^{T-1}$. Note that the policy depends on the current environment states.

We denote $\gamma \in [0,1]$ as the discounting factor. We use $V^{\boldsymbol{\pi}}_{i,t}: \mathcal{S}\rightarrow \mathbb{R}$ to denote the value function of agent $i$ at time index $t$ under guidance of the joint policy $\boldsymbol{\pi} = (\pi_1, \pi_2, \ldots, \pi_n)$, where $V^{\boldsymbol{\pi}}_{i,t}(s)
= \mathbb{E}_{\boldsymbol{\pi}} \left[\sum_{k=t}^{T-1} \gamma^{k-t}R_{i,k}(s_k,\mathbf{a}_k)\mid s_t=s \right]$. 
The Q-function  $Q^{\boldsymbol{\pi}}_{i,t}: \mathcal{S}\times\mathcal{A}\rightarrow\mathbb{R}$ of agent $i$ at time index $t$ with a state-joint action pair $(s,\mathbf{a})$ is defined to be $Q^{\boldsymbol{\pi}}_{i,t}(s,\mathbf{a}) = \mathbb{E}_{\boldsymbol{\pi}}\left[\sum_{k=t}^{T-1} \gamma^{k-t}R_{i,k}(s_k,\mathbf{a}_k)\mid s_t=s, \mathbf{a}_t=\mathbf{a}\right]$.
The relation between Q-value and V-value is $Q^{\boldsymbol{\pi}}_{i,t}(s,\mathbf{a}) = R_{i,t}(s, \mathbf{a}) + \gamma \sum_{s' \in \mathcal{S}}P_{t\rightarrow t+1}(s' \mid s,\mathbf{a})V^{\boldsymbol{\pi}}_{i,t+1}(s')$.

\section{MDP model for AI offloading and skill erosion}
\label{sec:AIoffloading}
 In this section, we consider the problem of balancing offloading tasks to the AIs and maintaining workers' skills.  We first introduce the MDP model.

\subsection{Model setup} 
We consider that workers have two possible skill levels: high levels and low levels. For each time step, the tasks (problems) the workers need to accomplish (solve) fall into one of the three types: training-enabled tasks, repeated tasks which do not enable training, and novel tasks.
Workers have two possible actions: either totally delegating to AI to work on the task or trying to finish the task by humans themselves (including humans using AIs as simple assisting tools).
Training tasks are the type of tasks (such as doing proof problems in calculus) that improve workers' skill while working on them, and thus the workers  have a higher probability of improving their skills if they work on the tasks independently.
Repeated tasks are relatively easy tasks (such as doing simple arithmetic operations) that offer a lower probability of skill improvement compared to training-type tasks.
Novel tasks are difficult tasks that would require novel never-seen-before solutions and fresh research data, and these tasks are challenging for both humans and AIs: however, high-skilled human workers can possibly solve them, and this results in high rewards for humans; while the AIs cannot finish them due to lack of training data or certain critical thinking skills and angles.    

Independent of worker' skill types, there are three possible types of problems or tasks to work on.  Thus, we structure states as a two-element tuple: the first element denotes the worker's skill level (high or low), and the 2nd element denotes the task type (training, repeat, novel) for the current time index.  
Thus, we have six states, labeled from $0$ to $5$.  The correspondence between state, skill level and task type is listed in Table \ref{tab:input table}.

For each state, the worker can choose between two possible actions: human efforts to solve or delegating to AI. The human action is labeled as $0$, while the AI action is labeled as $1$.

We define a probability transition kernel $P \in \mathbb{R}^{6 \times 2 \times 6}$ for this MDP, which represents the probabilities of the MDP transitioning between different states, under the two different actions.
$P[i,j,:]$ is a 6-dimensional probability vector representing the transition probabilities to each of the 6 next states, where $i$ ($0\leq i \leq 5$) denotes the current state and $j$ ($0\leq j\leq 1$) denotes the current action.  Let $p$, $q$ and $r$ be the probabilities of encountering each task type (training, repeat and novel, respectively) at each time index; and $p+q+r=1$. We assume that the task types are independent of the worker's skill level, and the task types are independent across time indices.  We also define a reward matrix $R \in \mathbb{R}^{6 \times 2}$, which represents the rewards the worker receives after working on the task. Here, $R(i,j)$ denotes the reward associated with state $i$ and action $j$. \\
For example, $P[0,0,:] = (0.7p, 0.7q, 0.7r, 0.3p, 0.3q, 0.3r)$ represents the transition probabilities to the $6$ different states, after a low-skilled worker works on a training-type problem using human efforts.  The worker has a probability of $0.7$ of remaining at the low-skill level and a probability of $0.3$ of transitioning to the high-skill level. Independent of the worker's skill level, the probabilities of encountering the three types of problems are determined by $p$, $q$, and $r$.

\begin{table}[h]
\centering
\resizebox{\textwidth}{!}{
\begin{tabular}{l l l l l l l l l l l}
\toprule
\multicolumn{2}{l}{State} & Action &Rate $\rho$ & \multicolumn{6}{l}{Transition kernel $P$} & $R$ Matrix \\
\cline{1-2}
\cline{5-10}
 state number &state detail & & & $p$ & $q$ & $r$ & $p$ & $q$ & $r$ & \\
\midrule
state 0 &Low, Training	& Human	&$\rho{(S_0,H)} = 0.7$		& 0.7p&0.7q&0.7r& 0.3p&0.3q&0.3r		&$R{(S_0,H)} = 10$ \\
state 0 &Low, Training	& AI 		&$\rho{(S_0,A)} = 1$			&  1p& 1q& 1r& 0p& 0q& 0r		&$R{(S_0,A)} = 15$ \\
state 1 &Low, Repeat	& Human&$\rho{(S_1,H)} = 0.95$ 		&  0.95p& 0.95q& 0.95r&0.05p& 0.05q& 0.05r		&$R{(S_1,H)} = 9$ \\
state 1 &Low, Repeat	& AI 		&$\rho{(S_1,A)} = 1$			&  1p& 1q& 1r& 0p& 0q& 0r 			&$R{(S_1,A)} = 14$ \\
state 2 &Low, Novel		& Human	&$\rho{(S_2,H)} = 0.8$  		&  0.8p& 0.8q& 0.8r& 0.2p& 0.2q& 0.2r 	&$R{(S_2,H)} = 0$ \\
state 2 &Low, Novel		& AI		&$\rho{(S_2,A)} = 1$  		&  1p& 1q& 1r& 0p& 0q& 0r 		&$R{(S_2,A)} = 1$ \\
state 3 &High, Training	& Human	&$\rho{(S_3,H)} = 0.01$  		&  0.01p& 0.01q& 0.01r&  0.99p& 0.99q& 0.99r		&$R{(S_3,H)} = 11$ \\
state 3 &High, Training	& AI		&$\rho{(S_3,A)} = 0.05$  		&  0.05p& 0.05q& 0.05r&  0.95p& 0.95q& 0.95r		&$R{(S_3,A)} = 16$ \\
state 4 &High, Repeat	& Human	&$\rho{(S_4,H)} = 0.03$  		&  0.03p& 0.03q& 0.03r&  0.97p& 0.97q& 0.97r	 	&$R{(S_4,H)} = 9.5$ \\
state 4 &High, Repeat	& AI		&$\rho{(S_4,A)} = 0.1$  		&  0.1p& 0.1q& 0.1r& 0.9p& 0.9q& 0.9r 		&$R{(S_4,A)} = 14.5$ \\
state 5 &High, Novel	& Human	&$\rho{(S_5,H)} = 0.001$  	&  0.001p& 0.001q& 0.001r& 0.999p& 0.999q&0.999r		&$R{(S_5,H)} = 100$ \\
state 5 &High, Novel	& AI		&$\rho{(S_5,A)} = 0.04$  		&  0.04p& 0.04q&0.04r& 0.96p& 0.96q& 0.96r			&$R{(S_5,A)} = 1$ \\
\bottomrule
\end{tabular}
}
\caption{The rate $\rho$ represents the probability that a worker deteriorates to or remains at the low-skill level after working on a task. We define $\rho \in [0,1]^{6 \times 2}$ as a probability matrix, where the first index represents the state and the second index represents the action chosen by the worker.}
% \vspace{0.5em}
% \small \textit{}
\label{tab:input table}
\end{table}
\FloatBarrier

We note the illustrative values of $\rho$ such as 0.7, 095, 0.8 etc. can be adjusted to reflect the actual values, depending on application scenarios.  The values of the rewards can also be adjusted.  

\subsection{Numerical results analysis}

\begin{figure}[ht]
    \centering
    \includegraphics[width=0.85\textwidth]{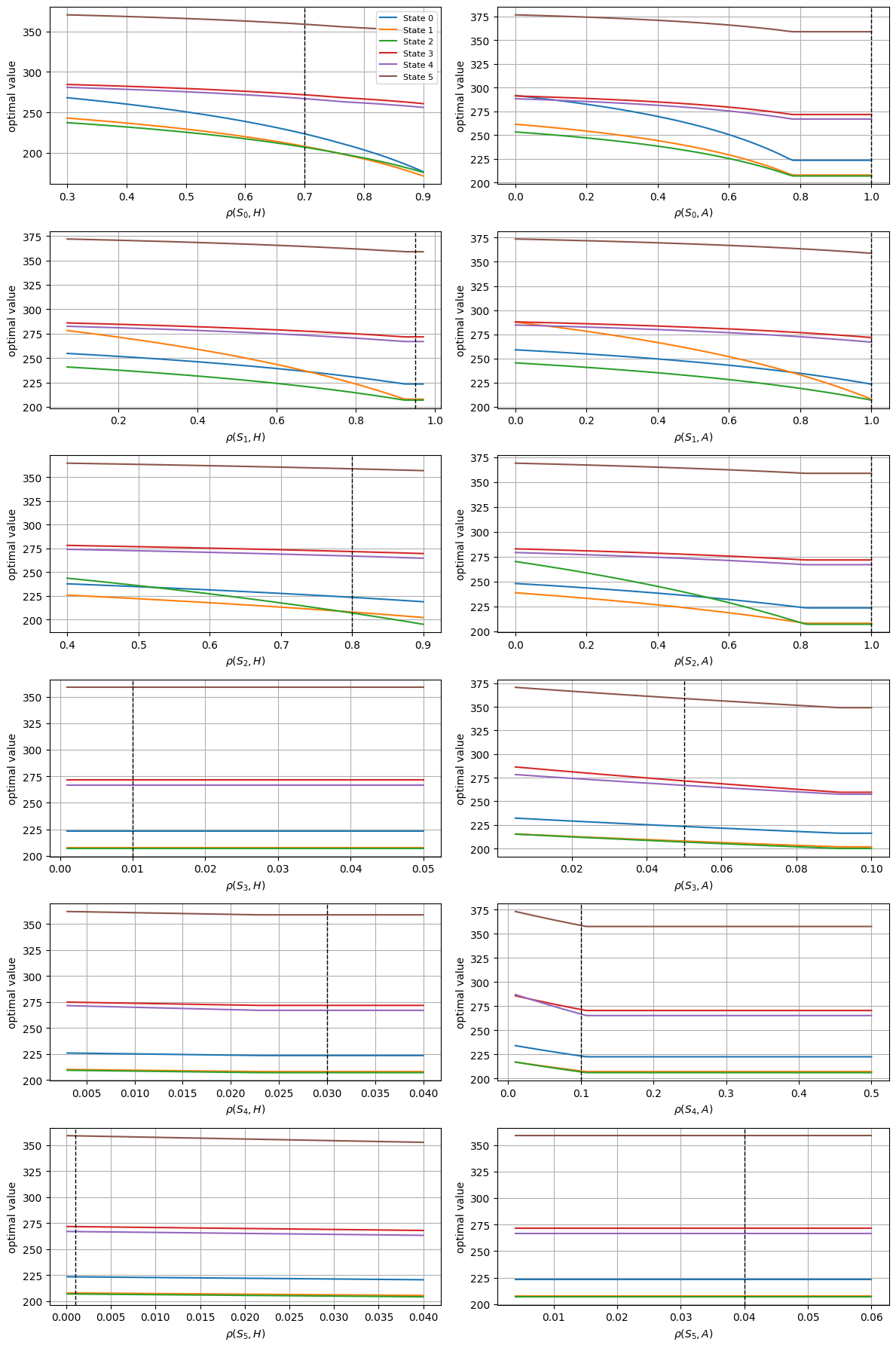}
    \caption{Optimal value vs $\rho$. This figure shows how the optimal value changes as each rate probability varies.}
    \label{fig:P_value}
\end{figure}

\begin{figure}[ht]
    \centering
    \includegraphics[width=0.85\textwidth]{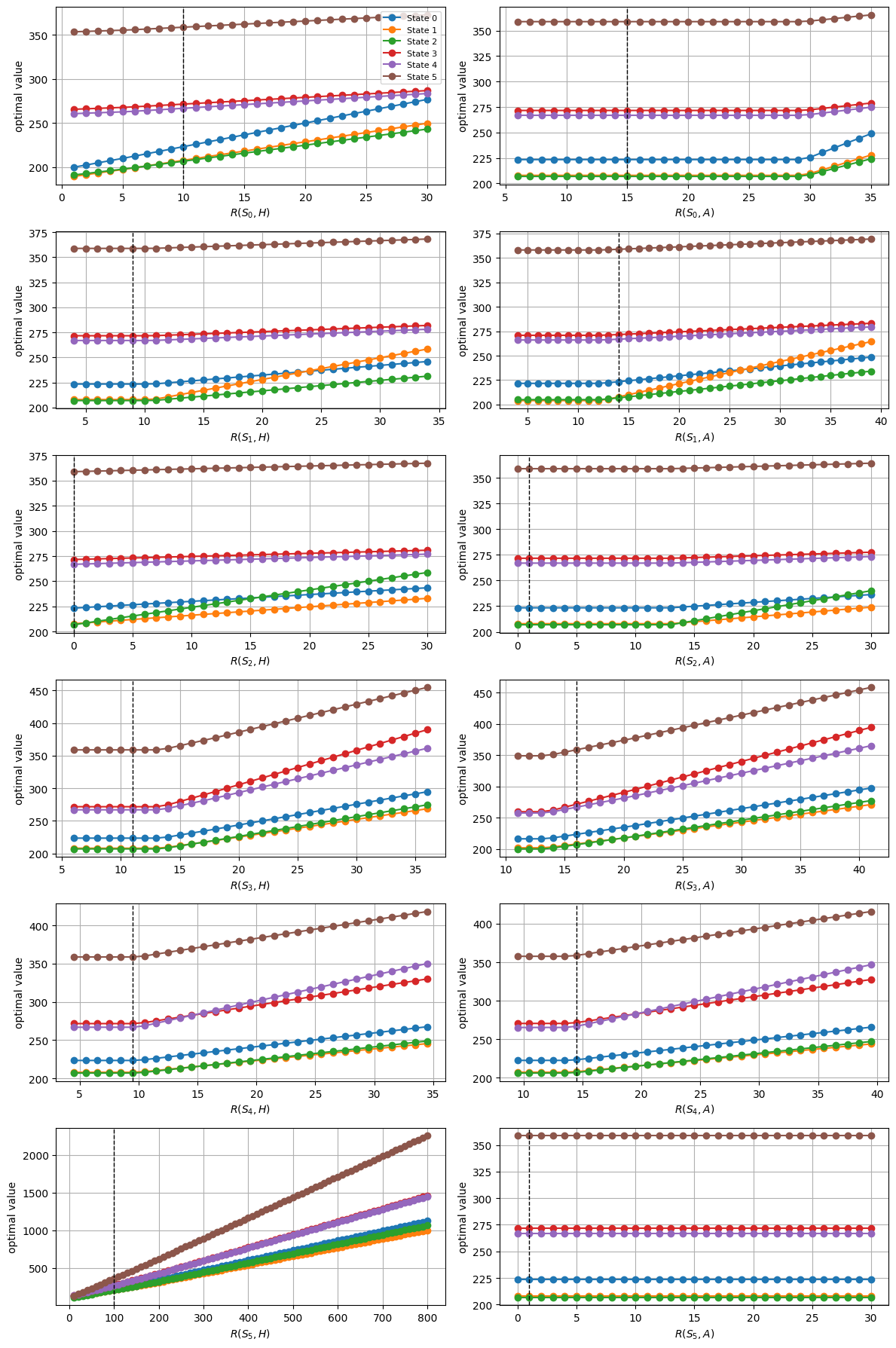}
    \caption{Optimal value vs $R$. This figure shows how the optimal value changes as each reward value varies.}
    \label{fig:R_value}
\end{figure}

\begin{figure}[ht]
    \centering
    \includegraphics[width=0.85\textwidth]{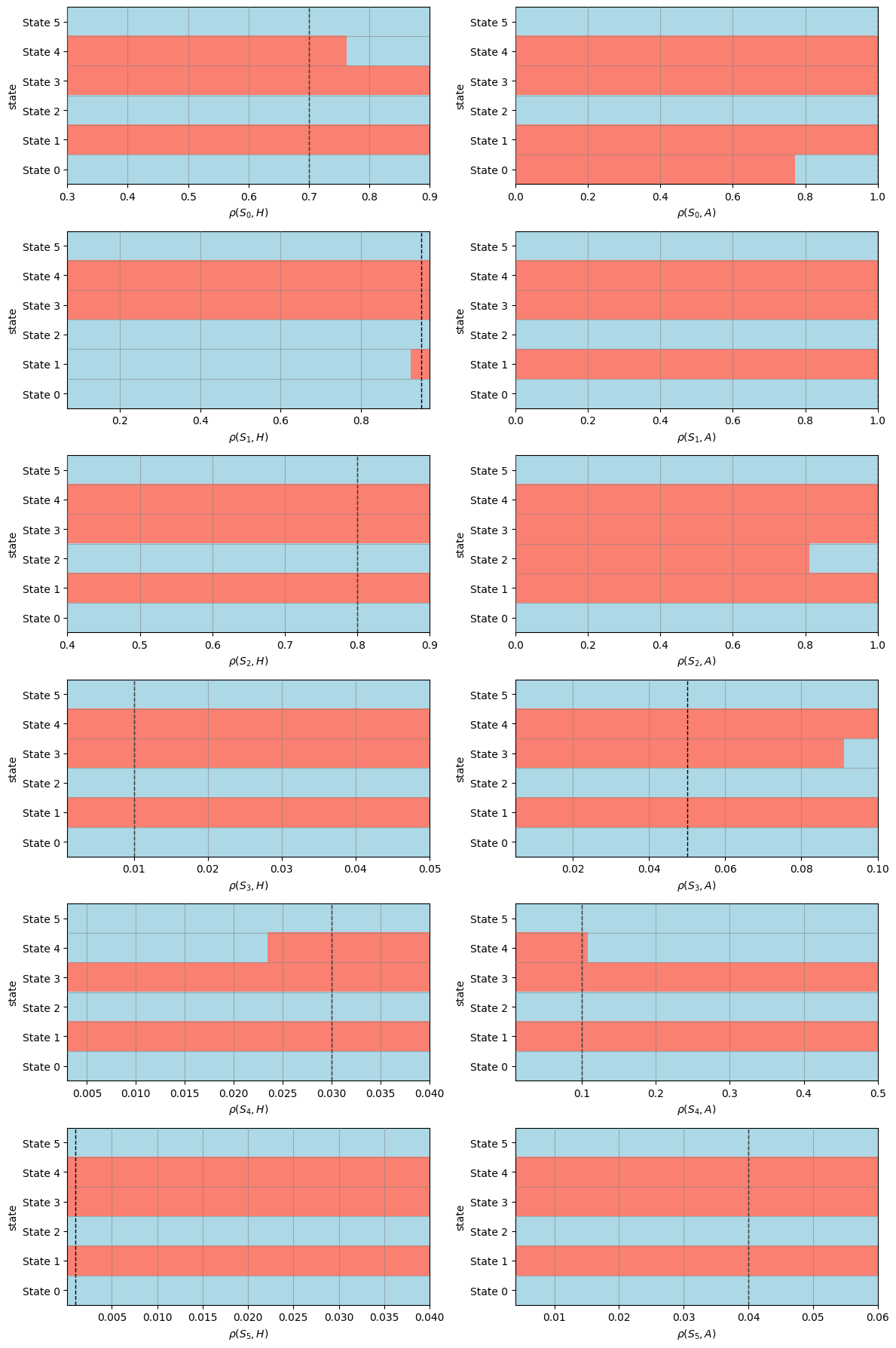}
    \caption{Optimal action vs $\rho$. This figure shows how the optimal action changes as the $\rho$ parameter changes. For each state, the 
    light blue color represents the worker choosing the human action, while salmon color represents the worker choosing the AI action.}
    \label{fig:P_action}
\end{figure}

\begin{figure}[ht]
    \centering
    \includegraphics[width=0.85\textwidth]{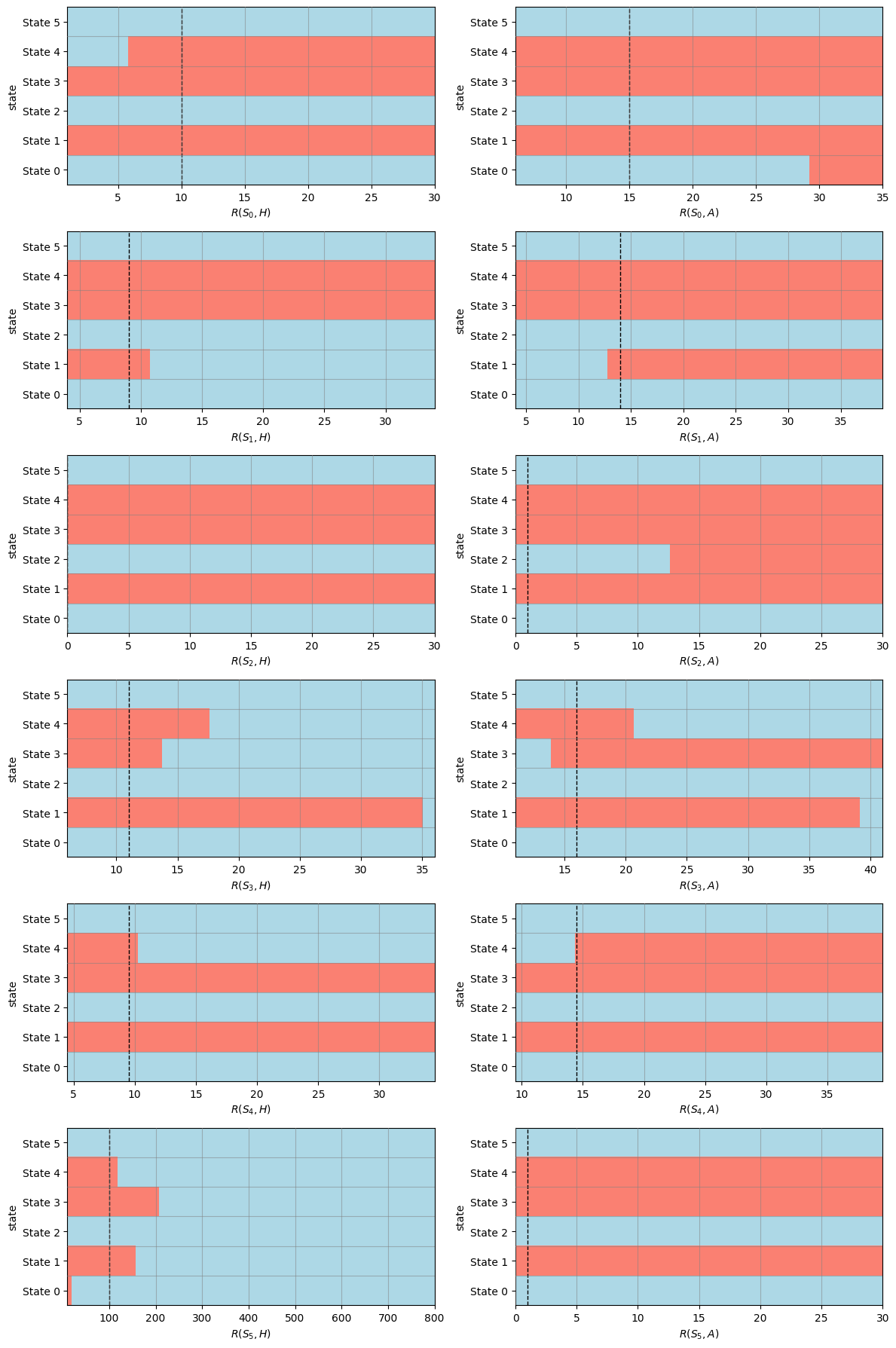}
    \caption{Optimal action vs $R$. This figure shows how the optimal action changes as each reward value varies. Light blue represents the worker choosing the human action, while salmon represents the worker choosing the AI action under each state.}
    \label{fig:R_action}
\end{figure}

We performed value iteration calculations for the MDP described in Table \ref{tab:input table}.  The results are reported in Figures \ref{fig:P_value}, \ref{fig:R_value},  \ref{fig:P_action}, and \ref{fig:R_action}. We set the discount factor to $\gamma = 0.9$.

In Figure~\ref{fig:P_value}, we plot the optimal V-values of each state as a function of the parameter $\rho$'s, where we vary the values of each $\rho$ parameter.  These plots show that the optimal V-values depend on the $\rho$ parameters, which govern how the use of AI will affect the worker's skill level.      

In Figure~\ref{fig:R_value}, we plot the  optimal V-values of each state as a function of the reward parameters $R$'s, where we vary the values of reward parameters. From this, we can see that the optimal V-values grow noticeably with $R(S_5, H)$, which is the reward the agent gets when working on a novel task with high human skills.

In Figure~\ref{fig:P_action}, by varying the parameters $\rho$'s, we observe how the best action for each state changes. We can see that when the agent is currently low-skilled and faced with a training type of task (state 0), the optimal action is actually human action, even though the immediate reward of AI action is higher than the human action.  We also see that when the agent is currently low-skilled, but faced with the ``repeat'' type of task (state 1), the agent will use the AI action since the AI action provides higher reward while not significantly reducing the chance of the worker becoming high-skilled compared with the human action.   We also observe for repeated tasks, the optimal actions are often the AI actions since the human actions offer lower immediate rewards while not increasing the probability of the worker becoming a high-skilled worker.

In Figure~\ref{fig:R_action}, we vary the values of the reward parameters, and we observe how the optimal action changes. We can see that when the worker is low-skilled and the agent is faced with a training type of task, the agent's optimal action is to use the human action for a large range of reward parameter values, even when the corresponding AI actions give more immediate rewards. In contrast, when the worker is already high-skilled (state 3), the worker will switch from human action to AI action if we decrease the reward $R(S_3, H)$ of human action or increase the reward $R(S_3, A)$ of AI actions. This is because the agent benefits less from improving the agent's skill level by working on the task by the human action, and is more willing to take the AI action.

\begin{figure}[ht]
    \centering
    \includegraphics[width=0.85\textwidth]{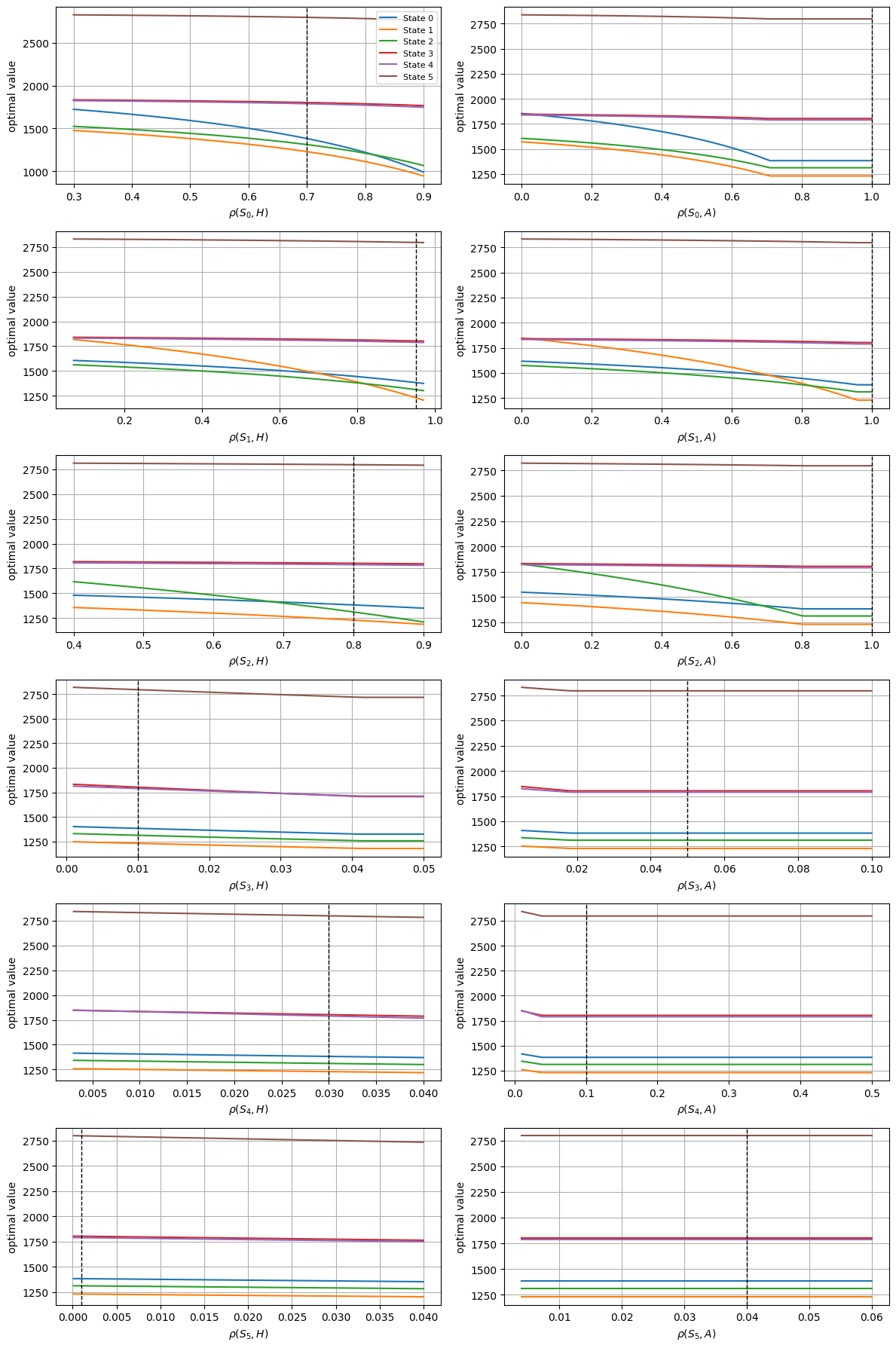}
    \caption{Optimal value vs $\rho$ with reward 1000.}
    \label{fig:P_val1000}
\end{figure}

\begin{figure}[ht]
    \centering
    \includegraphics[width=0.85\textwidth]{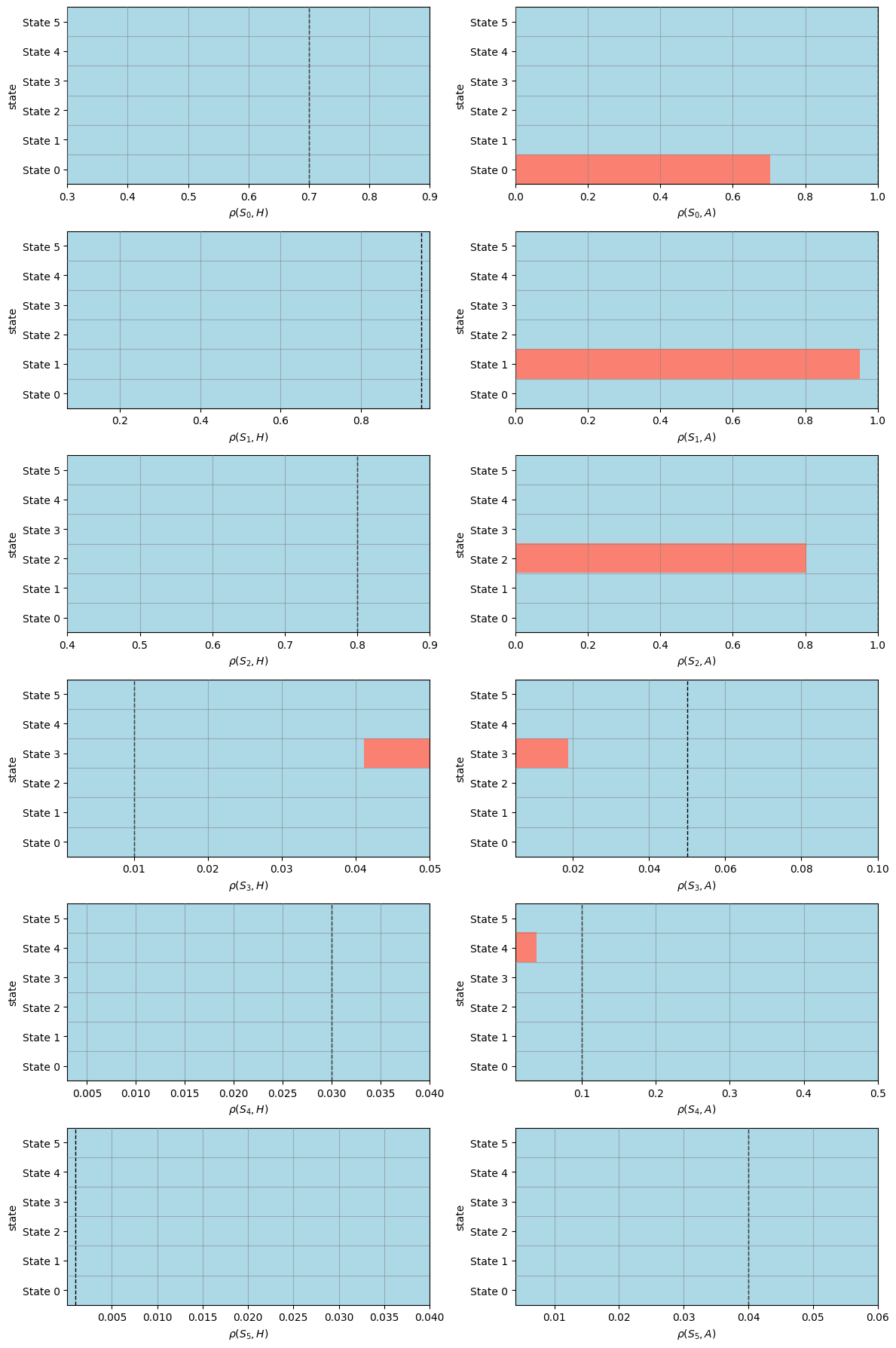}
    \caption{Optimal action vs $\rho$ with reward 1000.}
    \label{fig:P_action1000}
\end{figure}

We further plot Figures \ref{fig:P_val1000} and \ref{fig:P_action1000}, corresponding to extensions of Figures \ref{fig:P_value} and \ref{fig:P_action} respectively, after increasing the reward $R(S_5,H)$ from 100 to 1000. This means that a high-skilled worker will get more rewards when facing a novel task, increasing the reward amount from $100$ to $1000$. This is reasonable when completing a novel task (such as making an invention) can bring significantly more rewards.  From these two figures, especially from Figure \ref{fig:P_action1000}, we can see that the workers are more inclined to adopt human actions instead of AI actions under different states: this is because the human actions can better preserve/train the worker's skills such that a skilled agent will get more rewards whenever they are faced with a novel task.


\FloatBarrier
\section{MDP model for a firm's AI adoption and human resources buildup}
\label{sec:workforcebuildup}

In this section, we consider the optimal way for a firm to adopt AI and build its workforce. On the surface, it is tempting to keep only senior workers, who can accomplish the tasks with the help of AI, thus saving the cost of hiring junior workers. However, our MDP model shows that it is important to jointly optimize using AI and hiring junior workers.

In this model, we consider an MDP with states represented by a 3-element vector $(S, J, R)$. Here $S$ means whether there are senior workers in the firm: if there are, $S=1$; otherwise, $S=0$. $J$ denotes whether there are junior workers in the firm: if there are, $J=1$; otherwise, $J=0$. $R$ represents whether the junior workers (if there are) are ready to perform the tasks that can be handled by senior workers: if they are ready, $R=1$; otherwise, $R=0$. We do not allow the state $(0,1,1)$ to appear, since in that case, that state can be changed to $(1,0, 0)$ immediately. So in total, there are $5$ possible states: $(1,0,0)$, $(1,1, 0)$, $(1,1,1)$, $(0,1,0)$, and $(0,0,0)$.

We assume that there are no senior workers on the job market that are already trained to have domain expertise specifically for this firm. So the firm can only hire junior workers from the job market; and moreover, in our model, we consider the firm only hiring but not firing workers. 

We use another 3-dimensional vector $(SA, JA, EA)$ to represent the firm's possible actions.  $SA$ represents whether the senior workers adopt AI or not: if they do, $SA=1$; otherwise, $SA=0$. $JA$ denotes whether the junior workers adopt AI or not: if they do, $JA=1$; otherwise, $JA=0$. $EA$ denotes whether the company perform hiring junior workers: if $EA=1$, it means the company performs hiring; if $EA=0$, it means the company does not perform hiring. If the company performs hiring, and the existing junior workers are not promoted to senior workers,  the hiring will not add new junior workers. Moreover, junior workers become senior workers if and only if senior workers leave and the junior workers are ready  and able to perform senior workers' tasks. 

In this model, we assume that, when there is open space for hiring junior workers (for example, original junior workers have taken senior workers' positions), a hiring action is successful with probability $0.91$ and fails with probability 0.09. Hiring itself carries a reward of $-2$.  If junior workers who have not improved to the level of senior workers use AI, they will get to the level of senior workers with probability 0.3 at the next time index; if junior workers who have not improved to the level of senior workers use AI, they will get to the level of senior workers with probability 0.35 instead.  If there are no senior workers, junior workers adopting AI will produce a reward of 8, while junior workers not adopting AI will produce reward 6. But keeping junior workers in the workforce will also reduce the reward by 3, due to the labor cost.  Senior workers leave workforce with probability $0.93$ within each time step. When senior workers adopt AI, the firm will produce reward $30$, no matter what junior workers do.  When senior workers do not adopt AI, they will produce a reward of 10. But we do not separately consider the labor cost of keeping senior workers, since we already deduct the cost from the reward (namely the reward after deduction is 30 or 10).  We also assume that when senior workers use AI, junior workers do not contribute to producing any reward since under the senior workers' supervision,  the AI can replace the labor contribution from the junior workers.  
We remark these numbers (0.91, 0.93, 30, 10, 3, 2) can be generalized to general variables which can take other values.

Based on these rules, we have the following transition probability and reward structure for the corresponding MDP.

\subsection{State (0,0,0)}
Suppose that the current state is $(0,0,0)$, namely there are no senior or junior workers in this company. 

If the action is $(SA,JA, 1)$, where $SA$ and $JA$ can be of any value,  the state will transition to $(0,1,0)$ with probability $0.91$, and transition to $(0,0, 0)$ with probability $0.09$. The reward for these transitions is $-2$. 

If the action is $(SA,JA,0)$, where $SA$ and $JA$ can be any value, the MDP state will transition to $(0,0,0)$ with probability $1$, and this transition will produce reward $0$.

\subsection{State (0,1,0)}
Suppose that the current state is $(0,1,0)$, namely there are only junior workers in this company. 

\subsubsection{Action (SA,1,1)}
If the action is $(SA, 1, 1)$, where $SA$ can be any value, {junior workers adopt AI and the firm performs hiring}. Senior worker may or may not adopt AI, depending on what value $SA$ takes.

Then the state will transition to $(0,1,0)$ with probability $0.7$, with reward $8-3-2$=3, where $-3$ is the deduction caused by the labor cost of junior workers, and $-2$ is the reward of performing hiring.

The state will transition to $(1,1,0)$ with probability $0.3 \times 0.91$ and with reward $8-3-2=3$, where $0.3$ is the probability of the junior workers getting to senior workers' skill level, and $0.91$ is the probability of making a successful hiring.

The state will transition to $(1,0,0)$ with probability $0.3 \times 0.09$ and with reward $8-3-2=3$, where $0.3$ is the probability of the junior workers getting to senior workers' skill level, and $0.09=1-0.91$ is the probability of making an unsuccessful hiring.

\subsubsection{Action (SA,1,0)}
If the action is $(SA, 1, 0)$, where $SA$ can be of  any value, {junior workers adopt AI, and the
firm does not perform hiring. Senior worker may or may not adopt AI depending on what value $SA$ takes}.

Then the state will transition to $(0,1,0)$ with probability $0.7$ and with reward $8-3=5$.

The state will transition to $(1,0,0)$ with probability $0.3$ and with reward $8-3=5$.

\subsubsection{Action (SA,0,1)}
If the action is $(SA, 0, 1)$, where $SA$ can be any value, {junior workers do not adopt AI, and the
firm performs hiring. Senior worker may or may not adopt AI depending on what value $SA$ takes. }.

Then the state will transition to $(0,1,0)$ with probability $0.65$ and with reward $6-3-2$=1, where $6-3=3$ is the reward junior workers produced when not adopting AI, $-2$ is the deduction caused by hiring,  and 0.65 is the probability that junior workers staying as junior workers do not adopt AI.

The state will transition to $(1,1,0)$ with probability $0.91*0.35$ and with reward $6-3-2=1$.

The state will transition to $(1,0,0)$ with probability $0.09*0.35$ and with reward $6-3-2=1$. 

\subsubsection{Action (SA,0,0)}
If the action is $(SA, 0, 0)$, where $SA$ can be of any value: {namely junior workers do not adopt AI, and the
firm does not perform hiring. Senior workers may or may not adopt AI depending on what value $SA$ takes.}

Then the state will transition to $(0,1,0)$ with probability $0.65$ and with reward $6-3=3$.

The state will transition to $(1, 0,0)$ with probability $0.35$ and with reward $6-3=3$.

\subsection{State (1,1,0)}

Suppose that the current state is $(1,1,0)$, namely there are both senior and junior workers in the firm, but the junior workers are not ready to perform the tasks that can only be performed by senior-level workers. 

\subsubsection{Action (1,1,1)}

If the action is $(1, 1, 1)$, both senior and junior workers adopt AI, and the firm performs hiring.  

Then the state will transition to $(1,0, 0)$ with probability $0.07*(0.3)*0.09$  (where 0.07 is the probability that senior workers leave the firm) and with reward $(30-3-2)$=25.

The state will transition to
$(1, 1, 0)$ with probability
$0.93*0.7+0.07*0.3*0.91$ (seniors stay and  juniors do not improve; or seniors leave, juniors improve and hiring is successful) and with reward (30-3-2)=25.

The state will transition to 
$(1,1, 1)$ with probability $0.93*0.3$ (senior does not leave and junior improves to senior level) and with reward $30-3-2$=25.

The state will transition to 
$(0,1,0)$ with probability 
$0.07*0.7$ (namely seniors leave, juniors do not improve) and with reward 30-3-2=25.

The state will transition to 
$(0,0,0)$ with probability $0$ and reward 30-3-2=25 (where $3$ is the cost of employing junior workers, here the reward 30 comes from senior workers working with AI tools; 2 is the cost of hiring).

\subsubsection{Action (1,1,0)}

If the action is $(1, 1, 0)$, both senior and junior workers adopt AI, and the firm does not perform hiring.  

Then the state will transition to $(1,0, 0)$ with probability (0.07)*(0.3)  (where 0.07 is the probability that senior workers leaving the firm) and with reward $(30-3)$=27.

The state will transition to 
$(1,1, 0)$ with probability $0.93*0.7$ (senior does not leave and junior do not improve to senior level) and with reward $30-3$=27; 

The state will transition to 
$(1,1, 1)$ with probability $0.93*0.3$ (seniors does not leave and junior improves to senior level) and with reward $30-3$=27; 

The state will transition to 
$(0,1,0)$ with probability 
$0.07*0.7$ (namely seniors leave, juniors do not improve) and with reward 30-3=27.

The state will transition to 
$(0,0,0)$ with probability $0$ and reward 30-3=27 (where $3$ is the cost of employing junior workers, here the reward 30 comes from senior workers working with AI tools;).

\subsubsection{Action (0,1,0)}

If the action is $(0, 1, 0)$, senior do not adopt AI and junior workers adopt AI, and the firm does not perform hiring.  

Then the state will transition to $(1,0, 0)$ with probability (0.07)*(0.3)  (where 0.07 is the probability that senior workers leave the firm. Seniors leave and juniors improve) and with reward $(10+8-3)$=15. (in this case, their rewards can add, seniors have 10 without AI, juniors have 5 with AI (after excluding the cost))

The state will transition to 
$(1,1, 0)$ with probability $0.93*0.7$ (senior  workers do not leave and junior workers do not improve ) and produce reward $10+5$=15; 

The state will transition to 
$(1,1, 1)$ with probability $0.93*0.3$ (senior workers do not leave and junior workers improve to senior-worker level) and produce reward $10+5$=15.

The state will transition to 
$(0,1,0)$ with probability 
$0.07*0.7$ (namely seniors leave, juniors do not improve) and produce reward 10+5=15.

The state will transition to 
$(0,0,0)$ with probability $0$ and reward 10+5=15.

\subsubsection{Action (1,0,0)}

If the action is $(1, 0, 0)$, senior workers adopt AI but junior workers do not adopt AI; and the firm does not perform hiring.  

Then the state will transition to $(1,0, 0)$ with probability (0.07)*(0.35)  (where 0.07 is the probability that senior workers leaving the firm) and with reward $(30-3)$=27.

The state will transition to 
$(1,1, 0)$ with probability $0.93*0.65$ (senior does not leave and junior do not improve to senior level) and produce reward $30-3$=27.

The state will transition to 
$(1,1, 1)$ with probability $0.93*0.35$ (senior workers do not leave and junior workers improve to senior level) and produce reward $30-3$=27.

The state will transition to 
$(0,1,0)$ with probability 
$0.07*0.65$ (namely seniors leave, juniors do not improve) and produce reward 30-3=27.

The state will transition to 
$(0,0,0)$ with probability $0$ and produce reward 30-3=27 (where $3$ is the cost of employing junior workers, here the reward 30 comes from senior workers working with AI tools).

\subsubsection{Action (0,1,1)} % this is one action that can be deleted

If the action is $(0, 1, 1)$, senior workers do not use AI and junior workers adopt AI; and the firm perform hiring.  

Then the state will transition to $(1,0, 0)$ with probability (0.07)*(0.3)*0.09  (where 0.07 is the probability that senior workers leaving the firm) and with reward $(10+5)$-2=13.

The state will transition to 
$(1,1, 0)$ with probability $0.93*0.7$+$0.07*0.3*0.91$ (senior does not leave and junior do not improve; or senior leaves, junior improves, and hiring successful ) and with reward $10$+5-2=13.

The state will transition to 
$(1,1, 1)$ with probability $0.93*0.3$ (seniors does not leave and junior improves to senior level) and with reward $10+5$-2=13.

The state will transition to 
$(0,1,0)$ with probability 
$0.07*0.7$ (namely seniors leave, juniors do not improve) and with reward 10+5-2=13.

The state will transition to 
$(0,0,0)$ with probability $0$ and reward 10+5-2=13.

\subsubsection{Action (1,0,1)}

If the action is $(1, 0, 1)$, senior workers  use AI and junior workers do not adopt AI, and the firm performs hiring.  

Then the state will transition to $(1,0, 0)$ with probability (0.07)*(0.35)*0.09  (where 0.07 is the probability that senior workers leaving the firm and the hiring is not successful) and with reward $30-3-2$=25.

The state will transition to $(1,1, 0)$ with probability $0.93*0.65$+$0.07*0.35*0.91$ (senior does not leave and junior do not improve; or senior leaves, junior improves, and hiring successful ) and with reward $30-3-2$=25.

The state will transition to 
$(1,1, 1)$ with probability $0.93*0.35$ (seniors does not leave and junior improves to senior level) and with reward $30-3-2$=25.

The state will transition to 
$(0,1,0)$ with probability 
$0.07*0.65$ (namely seniors leave, juniors do not improve) and with reward $30-3-2$=25.
%$0.07*0.7$, per Jitong's suggestion, we change it to $0.07*0.65$

The state will transition to 
$(0,0,0)$ with probability $0$ and reward $30-3-2$=25.

\subsection{State (1,1,1)}

Suppose that the current state is $(1,1,1)$, namely there are both  junior and senior workers in this company; and junior workers are ready to replace senior workers if needed. 

\subsubsection{Action (1,1,1)}
If the action is $(1, 1, 1)$, senior workers  use AI and junior workers adopt AI, and the firm performs hiring.  

Then the state will transition to $(1,0, 0)$ with probability (0.07)*0.09  (where 0.07 is the probability that senior workers leaving the firm and the hiring is not successful) and with reward $30-3-2$=25.

The state will transition to $(1,1, 0)$ with probability $0.07*0.91$ (senior leaves and hiring successful ) and with reward $30-3-2$=25.

The state will transition to 
$(1,1, 1)$ with probability $0.93$ (seniors does not leave) and with reward $30-3-2$=25.

The state will transition to 
$(0,1,0)$ with probability 
$0$ (namely seniors leave, juniors do not improve) and with reward $30-3-2$=25.

The state will transition to 
$(0,0,0)$ with probability $0$ and reward $30-3-2$=25.

\subsubsection{Action (1,1,0)}
If the action is $(1, 1, 0)$, senior workers  use AI and junior workers adopt AI, and the firm does not perform hiring.  

Then the state will transition to $(1,0, 0)$ with probability (0.07)  (where 0.07 is the probability that senior workers leaving the firm and junior workers are promoted) and with reward $30-3$=27.

The state will transition to $(1,1, 0)$ with probability 0  and with reward $30-3$=27.

The state will transition to 
$(1,1, 1)$ with probability $0.93$ (seniors do not leave) and with reward $30-3$=27.

The state will transition to 
$(0,1,0)$ with probability 
$0$ (namely seniors leave, juniors do not improve) and with reward $30-3$=27.

The state will transition to 
$(0,0,0)$ with probability $0$ and reward $30-3$=27.

\subsubsection{Action (0,1,0)}

If the action is $(0, 1, 0)$, senior workers do not use AI and junior workers adopt AI, and the firm does not perform hiring.  

Then the state will transition to $(1,0, 0)$ with probability (0.07)  and with reward $10+5$=15.

The state will transition to $(1,1, 0)$ with probability $0$  and with reward $10+5$=15.

The state will transition to 
$(1,1, 1)$ with probability $0.93$  and with reward $10+5$=15,

The state will transition to 
$(0,1,0)$ with probability 
$0$ (namely juniors do not improve) and with reward $15$.

The state will transition to 
$(0,0,0)$ with probability $0$ and reward $15$.

\subsubsection{Action (1,0,0)}

If the action is $(1, 0, 0)$, senior workers  use AI and junior workers do not adopt AI, and the firm does not perform hiring.  

Then the state will transition to $(1,0, 0)$ with probability (0.07)  (where 0.07 is the probability that senior workers leaving the firm) and with reward $30-3$=27.

The state will transition to $(1,1, 0)$ with probability $0$  and with reward $30-3$=27.

The state will transition to 
$(1,1, 1)$ with probability $0.93$ (seniors does not leave  and with reward $30-3$=27.

The state will transition to 
$(0,1,0)$ with probability 
$0$ reward $30-3$=27.

The state will transition to 
$(0,0,0)$ with probability $0$ and reward $30-3$=27.

\subsubsection{Action (0,1,1)}

If the action is $(0, 1, 1)$, senior workers do not  use AI, junior workers adopt AI, and the firm performs hiring.  (Note that in this case, junior workers using AI are the same as senior workers adopting AI)

Then the state will transition to $(1,0, 0)$ with probability (0.07)*0.09  (where 0.07 is the probability that senior workers leaving the firm and the hiring is not successful) and with reward $10+5-2$=13.

The state will transition to $(1,1, 0)$ with probability $0.07*0.91$ (or senior leaves and and hiring successful ) and with reward $10+5-2$=13.

The state will transition to 
$(1,1, 1)$ with probability $0.93$ (seniors does not leave) and with reward $10+5-2$=13.

The state will transition to 
$(0,1,0)$ with probability 
$0$ (namely seniors leave, juniors do not improve) and with reward $10+5-2$=13.

The state will transition to 
$(0,0,0)$ with probability $0$ and reward $10+5-2$=13.

\subsubsection{Action (1,0,1)}

If the action is $(1, 0, 1)$, senior workers  use AI and junior workers do not adopt AI, and the firm performs hiring.  

Then the state will transition to $(1,0, 0)$ with probability 0.07*0.09, $30-3-2$=25.

The state will transition to $(1,1, 0)$ with probability $0.07*0.91$ (senior leaves  and hiring successful ) and with reward $30-3-2$=25.

The state will transition to 
$(1,1, 1)$ with probability $0.93$ (seniors does not leave) and with reward $30-3-2$=25.

The state will transition to 
$(0,1,0)$ with probability 
$0$  and with reward $30-3-2$=25.

The state will transition to 
$(0,0,0)$ with probability $0$ and reward $30-3-2$=25.

\subsection{State (1,0,0)}

Suppose that the current state is $(1,0,0)$, namely there are only senior workers in this company. 

\subsubsection{Action (1,1,1)}
If the action is $(1, 1, 1)$, senior workers use AI and junior workers adopt AI, and the firm performs hiring.  

Then the state will transition to $(1,0, 0)$ with probability 0.93*0.09  (where 0.93 is the probability that senior workers not leaving the firm and the hiring is not successful) and with reward $30-2$=28.

The state will transition to $(1,1, 0)$ with probability $0.93*0.91$ (senior not leaving and hiring successful ) and with reward $30-2$=28.

The state will transition to 
$(1,1, 1)$ with probability $0$  and with reward $30-2$=28.

The state will transition to 
$(0,1,0)$ with probability 
$0.07*0.91$ (namely seniors leave, hiring successful) and with reward $30-2$=28.

The state will transition to 
$(0,0,0)$ with probability $0.07*0.09$ and reward $30-2$=28.

\subsubsection{Action (1,1,0)}

If the action is $(1, 1, 0)$, senior workers  use AI and junior workers do adopt AI (however there are no junior workers), and the firm does not perform hiring.  

Then the state will transition to $(1,0, 0)$ with probability (0.93)  (where 0.07 is the probability that senior workers leaving the firm) and with reward $30$.

The state will transition to $(1,1, 0)$ with probability $0$ and with reward $30$.

The state will transition to 
$(1,1, 1)$ with probability $0$ (seniors do not leave and junior improves to senior level) and with reward $30$.

The state will transition to 
$(0,1,0)$ with probability 
$0$  and with reward $30$.

The state will transition to 
$(0,0,0)$ with probability $0.07$ and reward $30$.

\subsubsection{Action (0,1,0)}

If the action is $(0, 1, 0)$, senior workers do not use AI and junior workers adopt AI, and the firm does not perform hiring.  

Then the state will transition to $(1,0, 0)$ with probability (0.93)  and with reward $10$ (10 is the reward when seniors do not use AI).

The state will transition to $(1,1, 0)$ with probability $0$  and with reward $10$.

The state will transition to 
$(1,1, 1)$ with probability $0$  and with reward $10$.

The state will transition to 
$(0,1,0)$ with probability 
$0$ (namely juniors do not improve) and with reward $10$.

The state will transition to 
$(0,0,0)$ with probability $0.07$ and reward $10$.

\subsubsection{Action (1,0,0)}

If the action is $(1, 0, 0)$, senior workers  use AI and junior workers do not adopt AI (there are no junior workers), and the firm does not perform hiring.  

Then the state will transition to $(1,0, 0)$ with probability (0.93)  (where 0.07 is the probability that senior workers leaving the firm) and with reward $30$.

The state will transition to $(1,1, 0)$ with probability $0$ and with reward $30$. 

The state will transition to 
$(1,1, 1)$ with probability $0$ (seniors do not leave and junior improves to senior level) and with reward $30$.

The state will transition to 
$(0,1,0)$ with probability 
$0$  and with reward $30$.

The state will transition to 
$(0,0,0)$ with probability $0.07$ and reward $30$.

\subsubsection{Action (0,1,1)}

If the action is $(0, 1, 1)$, senior workers does not  use AI and junior workers adopt AI, and the firm performs hiring.  

Then the state will transition to $(1,0, 0)$ with probability (0.93)*0.09  and with reward $10-2=8$.

The state will transition to $(1,1, 0)$ with probability $0.93*0.91$  and with reward $10-2$=8.

The state will transition to 
$(1,1, 1)$ with probability $0$ (seniors does not leave) and with reward $10-2$=8.

The state will transition to 
$(0,1,0)$ with probability 
$0.07*0.91$ (namely seniors leave and hiring is successful) and with reward $10-2$=8.

The state will transition to 
$(0,0,0)$ with probability $0.07*0.09$ and reward $10-2$=8.

\subsubsection{Action (1,0,1)}
If the action is $(1, 0, 1)$, senior workers  use AI and junior workers do not adopt AI, and the firm performs hiring. 

Then the state will transition to $(1,0, 0)$ with probability $0.93*0.09$ (where 0.93 is the probability that senior workers not leaving the firm and the hiring is not successful) and with reward $30-2$=28.

The state will transition to $(1,1, 0)$ with probability $0.93*0.91$ (senior not leaving and hiring is successful) and with reward $30-2$=28.

The state will transition to 
$(1,1, 1)$ with probability $0$  and with reward $30-2$=28.

The state will transition to 
$(0,1,0)$ with probability 
$0.07*0.91$ (namely seniors leave, hiring successful) and with reward $30-2$=28.

The state will transition to 
$(0,0,0)$ with probability $0.07*0.09$ and reward $30-2$=28.

\subsection{Numerical result analysis}
For notational convenience, we represent each state using a vector notation as discussed before. Specifically, $S_0=(0,0,0)$, $S_1=(0,1,0)$, $S_2=(1,0,0)$, $S_3=(1,1,0)$, $S_4=(1,1,1)$.
Similarly, we denote each action using a vector notation, where $A_0=(1,0,0)$, $A_1=(1,1,0)$,  $A_2=(1,0,1)$, $A_3=(0,1,1)$, $A_4=(1,1,1)$, and $A_5=(0,1,0)$.

\begin{figure}[ht]
    \centering
    \includegraphics[width=0.85\textwidth]{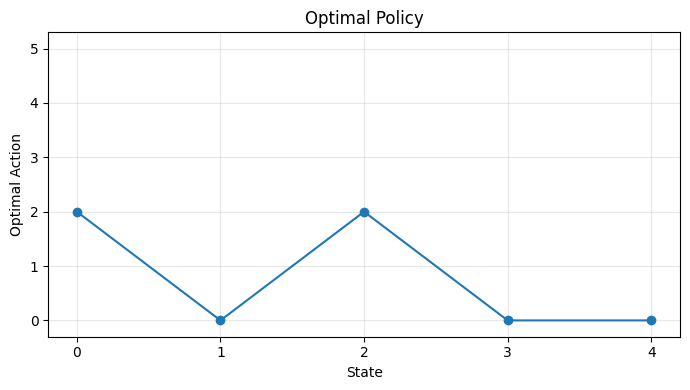}
    \caption{Optimal action vs state. This figure shows what the optimal action is for every state. Here number $i$ for action corresponds to action $A_{i}$.}
    \label{fig:bestaction}
\end{figure}

In Figure~\ref{fig:bestaction}, we plot the best action for each state. Here, Action $i$ means Action $S_{i}$. From this figure, we can see that, under states $S_0$ and $S_2$, it is optimal to select action $A_2$, while for all the remaining states, it is optimal to select action $A_0$. States $S_0$ and $S_2$ represent situations in which the firm currently has no junior workers. Under these states, in the optimal action $A_2$, senior workers adopt AI and the firm hires new junior workers. Even though AI can enable senior workers to generate high rewards alone in the absence of junior workers, the optimal action is still to hire junior workers and keep them as reserved workforce even though junior workers do not contribute to increasing the reward at this moment.    

In contrast, the remaining states represent situations in which junior workers are present in the firm. In these states, the preferred action is $A_0$, where junior workers do not use AI offloading, senior workers adopt AI, and the firm does not hire additional workers. This allows junior workers to develop their skills through human efforts,  and these trainings provide them with opportunities to become senior workers in the future.  If the junior workers use offloading, their skill level can grow more slowly and thus cannot become senior workers in time to generate higher rewards.   

\FloatBarrier
\section{MDP models for economics of AI Agents}
\label{sec:economics}
We consider a trading game between $n$ agents, which may include both humans and AI agents.  

\subsection{Multi-agent economic exchange problem}
We consider $n$ agents, where $n$ is a positive integer. Each agent can function as a producer and also as a consumer. Note that an ``agent'' in our model may represent a human agent, an AI agent, or even an entity of collaborative AI/human agents (for example, the collection of all the people in a country can be modeled as an agent in international trade; the collection of all the AI agents owned and controlled by a technology company can be modeled a single ``big'' agent in providing products/services). 
We consider a spending matrix $P\in \mathbb{R}^{n\times n}$, where $P$ is a Markov chain transition ``probability'' matrix. We denote the element in the $i$-th row and $j$-th column of $P$ as $P_{ij}$. For matrix $P$, we have
$\sum_{i=1}^{n}P_{ij}=1$ for every $1 \leq j \leq n$. The number $P_{ij}$ denotes the fraction of the $j$-th player's currency it spends on purchasing the products/services from the $i$-th agent. Note that the numbers in the Markov chain transition probability matrix do not represent probabilities in their physical meanings: they just denote the fraction of one agent's money spent on purchasing products/services from another agent.  

We use $x^t \in \mathbb{R}^{n\times 1}$ to denote the currency vector at time index $t$, where $t\geq 0$ is an integer.  We assume that each element of $x^t$ is non-negative, representing the currency possessed by each agent at time index $t$. We refer to the time interval between consecutive time indices as one episode.  We  have a utility matrix  $U\in \mathbb{R}^{n\times n}$ whose elements are non-negative. We denote the element in the $i$-th row and the $j$-th column of $U$ as $U_{ij}$, which represents the utility the $i$-th player provides to the $j$-th agent if the $j$-th agent spends one dollar purchasing products/services from agent $i$.  % We assume that no agent changes their spending pattern as time evolves, namely a player maintains the relative ratios of spending on each player.

 The goal of each agent is to maximize its obtained utility.  On the surface, to maximize the utility of the $j$-th agent, the $j$-th agent should spend all its currency on the agent who provides the highest utility for each dollar spent on it, namely the agent $i$ with the largest $U_{ij}$. However, counter-intuitively we argue that this is not true, and the optimal spending practice should depend on the structure of the network of producers and consumers. We notice that at the time index $(t+1)$,  the currency vector is updated to 
$$ x^{t+1}=Px^{t}.$$

We can also let the spending matrix $P$ vary as functions of time $t$, and we denote it as $P^t$. The spending matrix $P^{t}$ model the actions of all the agents: its $j$-th column is the $j$-agent's actions at time index $t$. The state of agent $i$ is described by the wealth possessed by that agent. 

In the following, we give one example which shows the multi-agent MDP model provides a solution to how to best integrate the AI agents in the economic game.

\subsection{An example of the economic trading game}
 We introduce an example of the economic game.  Suppose that there are $n=3$ agent: two of them are human agents $H_1$ and $H_2$; the other one is an AI agent. 

 Suppose that there is only one unit of total wealth in the system. Suppose $\gamma<1$ is the discounting factor.  Agent $H_1$ has two possible actions when it has one unit of wealth.  The first action is to purchase services from the AI agent, when $H_1$ gives one unit of wealth to the AI agent and obtains environmental reward (namely utility) $100$; the 2nd action is to purchase services from $H_2$, when $H_1$ gives one unit of wealth to $H_2$ and $H_1$ obtains environmental reward $5$. Agent $H_2$ has two possible actions when it has one unit of wealth.  The first action is to purchase services from the AI agent, when $H_2$ gives one unit of wealth to the AI agent and obtains environmental reward $80$; the 2nd action is to purchase services from $H_1$, when $H_2$ gives one unit of wealth to $H_1$ and agent $H_2$ obtains environmental reward $100$.
The AI agent has three actions when it has one unit of wealth: the first action is to do the job itself, getting environmental reward of 1000; the second action is to purchase its services from $H_1$ at the cost of 1 unit of wealth, obtaining an environmental reward of $30$; the third action is to purchase its services from $H_2$ at the cost of 1 unit of wealth, obtaining an environmental reward of $30$. When an agent does not have any wealth, they cannot purchase services from others or perform the jobs itself. 

Then the following policies for the three agents form an equilibrium: Agent $H_1$ always purchases from Agent $H_2$ whenever it has one unit of wealth; Agent $H_2$ always purchases from Agent $H_1$ whenever it has one unit of wealth; and the AI agent always purchases from itself whenever it has one unit of wealth.     
   
To see this is an equilibrium, we show that no one can improve their own optimal value by changing their policy while others keep their policies.  First of all, we notice that it is always optimal for the AI agent to purchase its own services since it achieves the largest possible reward 1000 for every time step.  Suppose that the optimal policy of Agent $H_1$ is to purchase the services of the AI agent whenever it has a unit of wealth. Then it will only have 100 discounted reward since later on the AI agent will spend the wealth on purchasing the services from itself. However, if Agent $H_1$ instead spends the unit wealth on purchasing the services from Agent $H_2$, the obtained discounted rewards will be $\frac{5}{1- \gamma^2}$. So if $\gamma$ is close to $1$ or we consider undiscounted reward, this reward will be bigger than the reward obtained through the $H_1$'s policy of purchasing from the AI agent. Similar conclusion also holds for Agent $H_2$. So indeed, it is a Nash equilibrium. We note that in this equilibrium the optimal policies of the human agents are not to purchase from the AI agent, even though the AI agent provides better instant reward.
 
For the equilibrium above, one may wonder why the optimal strategy of a human agent is not to purchase from the most powerful AI agent who provides the highest utility per dollar spent on it.  This is because in our model, the spending pattern or strategy of an agent not only affects its instant utility obtained from spending, but also affects the wealth state the agent is in.

\section{Conclusions and Future Directions}
\label{sec: conclusions}

In this paper, we develop a general multi-agent Markov decision process (MDP) based reinforcement learning (RL) framework for determining and analyzing optimal adoption of AI technologies.  This framework can model the long-term dynamic effects of AI adoption, and lead to finding the optimal AI adoption strategies. In our modeling framework, the negative effects of AI adoption are modeled by state transitions or negative rewards in the MDPs. The proposed multi-agent RL incorporates both human agents and AI agents.

 The key idea of our models is that incorporating AI technologies and AI agents may not only increase the immediate rewards for AI agents but also alter the states of the involved human and AI agents. In economy, adopting AI may affect the job opportunity and wealth distributions among agents, resulting in different wealth states or job opportunity states the agents are in.  For a firm, using AI technologies to replace junior workers can result in unhealthy workforce preparation when faced with loss of senior workers or when faced with novel types of future challenges. 
 
 In future work, it is important to explore more general policies like history-dependent mutli-agent RL policies in adopting AI technologies \cite{lu2026historymattersmetapolicydelegation}.

\clearpage
\begin{spacing}{1.6}
\typeout{[Bibliography]}
\bibliography{Reference_20260227,Reference_NullSpaceCondition}
\end{spacing}
\end{document}